\documentclass[letterpaper]{article} 
\usepackage{aaai2026}  
\usepackage{times}  
\usepackage{helvet}  
\usepackage{courier}  
\usepackage[hyphens]{url}  
\usepackage{graphicx} 
\usepackage{natbib}  
\usepackage{caption} 
\usepackage{algorithm}
\usepackage{algorithmic}

\usepackage{amssymb}
\usepackage{booktabs}
\usepackage{float}
\usepackage{tabularx}
\usepackage{bm}
\usepackage{mathrsfs}
\usepackage{amsmath}
\usepackage{xcolor} 
\usepackage{pifont}
\usepackage{newfloat}
\usepackage{listings}
\DeclareCaptionStyle{ruled}{labelfont=normalfont,labelsep=colon,strut=off} 
\floatstyle{ruled}
\newfloat{listing}{tb}{lst}{}
\floatname{listing}{Listing}
\title{PersonaAnimator: Personalized Motion Transfer from Unconstrained Videos}
\author{
    Ziyun Qian\textsuperscript{\rm 1}\equalcontrib,
    Runyu Xiao\textsuperscript{\rm 1}\equalcontrib,
    Shuyuan Tu\textsuperscript{\rm 2},
    Wei Xue\textsuperscript{\rm 1},
    Dingkang Yang\textsuperscript{\rm 1}\thanks{Corresponding authors.},
    Mingcheng Li\textsuperscript{\rm 1},
    Dongliang Kou\textsuperscript{\rm 1},
    Minghao Han\textsuperscript{\rm 1},
    Zizhi Chen\textsuperscript{\rm 1},
    Lihua Zhang\textsuperscript{\rm 1}\thanks{Corresponding authors.}
}
\affiliations{
    \textsuperscript{\rm 1}College of Intelligent Robotics and Advanced Manufacturing, Fudan University\\
    \textsuperscript{\rm 2}College of Computer Science and Artificial Intelligence, Fudan University\\

    zyqian22@m.fudan.edu.cn

}

\usepackage{bibentry}

\begin{document}

\maketitle

\begin{figure*}[t]
  \centering
  \includegraphics[width=\textwidth]{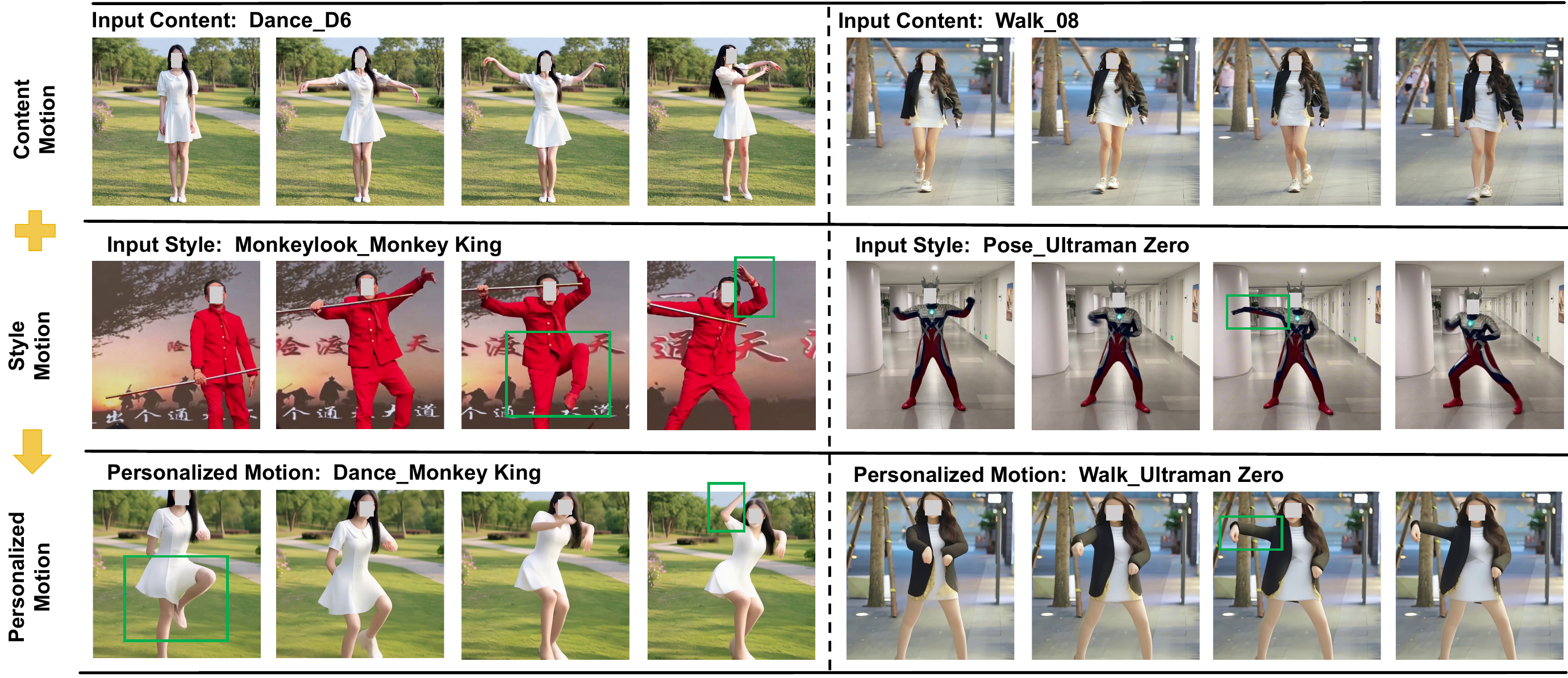}
    \captionof{figure}{Video-to-Video Motion Personalization. Our method preserves the basic motion content while learning personalized motion characteristics from the style motion video, achieving personalized transfer. Learned personalized features are marked with green bounding boxes.}
    \label{引言图}
\end{figure*}

\begin{abstract}
Recent advances in motion generation show remarkable progress. However, several limitations remain: (1) Existing pose-guided character motion transfer methods merely replicate motion without learning its style characteristics, resulting in inexpressive characters. (2) Motion style transfer methods rely heavily on motion capture data, which is difficult to obtain. (3) Generated motions sometimes violate physical laws. To address these challenges, this paper pioneers a new task: Video-to-Video Motion Personalization. We propose a novel framework, PersonaAnimator, which learns personalized motion patterns directly from unconstrained videos. This enables personalized motion transfer. To support this task, we introduce PersonaVid, the first video-based personalized motion dataset. It contains 20 motion content categories and 120 motion style categories. We further propose a Physics-aware Motion Style Regularization mechanism to enforce physical plausibility in the generated motions. Extensive experiments show that PersonaAnimator outperforms state-of-the-art motion transfer methods and sets a new benchmark for the Video-to-Video Motion Personalization task.
\end{abstract}


\section{Introduction}

With the rapid rise of the metaverse and digital content creation, building digital avatars with a strong sense of identity has become a central requirement for interactive experiences. Motion style, as a highly distinctive representation of individual behavior, not only captures subtle kinematic variations but also conveys high-level semantic cues such as identity and emotion—for instance, their gait or signature gestures can often recognize a person. \cite{style_def1,style_def2,gait}
In computer vision, prior works \cite{zhao2024dartcontrol,18-wang2025humandreamer} have achieved basic motion generation, but key limitations remain:
(1) Pose-guided character motion transfer methods \cite{17-tu2024motioneditor,16-li2024dispose,wang2024unianimate} merely mechanically replicate motion without capturing personalized traits, severely limiting the expressiveness of digital characters in real-world applications.
(2) Motion style transfer techniques \cite{19-aberman2020unpaired,22-kim2024most,23-kim2025personabooth} can learn style representations. However, they rely wholly or partially on motion capture data, which is costly, labor-intensive, and often unavailable for historical figures or virtual characters.
(3) Existing methods generally lack explicit physical constraints on the full-body skeleton, leading to distorted or implausible motion.

To address the above challenges, we propose the novel task of Video-to-Video Motion Personalization, which firstly enables personalized motion transfer from unconstrained videos, as illustrated in Fig. \ref{引言图}. Our PersonaAnimator framework makes two key breakthroughs:
First, it is the first to learn personalized motion features entirely from ordinary video data, without relying on any motion capture data, making it feasible to model historical figures or virtual characters for whom such data is unavailable.
Second, our framework preserves the structural integrity of the content motion while learning personalized style characteristics, generating personalized motions. Rather than performing simple, mechanical motion transfer, it establishes a new paradigm that shifts from ``postural approximation'' to ``personalized dynamic essence'' in motion generation.
These innovations enable wide applications in digital content creation—for example, allowing stunt doubles to replicate iconic celebrity movements (\emph{e.g.}, Michael Jackson’s ``Moonwalk'') or letting users embed signature gestures (\emph{e.g.}, a chin rest) into virtual avatars, offering critical support for building digital humans with authentic identities.

To train the video-to-video motion personalization model, we introduce PersonaVid, the first video-based personalized motion dataset, comprising 20 motion content categories and 120 motion style categories. We adopt a “one person, one style” paradigm, treating each individual as a unique style class to better capture personalized motion traits.
Additionally, we propose the Physics-aware Motion Style Regularization (PMSR) mechanism, which enforces physically plausible motion through dynamic bone stability and body connectivity constraints.
Extensive experiments demonstrate that PersonaAnimator achieves state-of-the-art (SOTA) results on both PersonaVid and mainstream human animation datasets \cite{tiktok_dataset,fashion_dataset}. The main contributions are summarized as follows:

\begin{itemize}
\item We pioneer a novel task, Video-to-Video Motion Personalization, and propose a new framework, PersonaAnimator, which incorporates the SA-PMT module to enable personalized motion transfer from unconstrained videos.
\item We introduce PersonaVid, the first video-based personalized motion dataset.
\item We propose the PMSR mechanism to enforce physical plausibility in generated motions.
\item Extensive experiments show that PersonaAnimator outperforms SOTA methods.
\end{itemize}

\section{Related Work}
\subsection{Diffusion Models for Video Generation}
Diffusion models are central to video generation, focusing on two directions: Text-to-Video (T2V) \cite{1-xing2024survey,2-ho2022video,3-ho2022imagenvideohighdefinition,4-lin2023videodirectorgpt,5-fei2023empowering} and unconditional generation \cite{6-mei2023vidm,7-lu2023vdt,8-fei2024vitron}.
VDM \cite{2-ho2022video} pioneered diffusion-based T2V by extending 2D U-Net to 3D. Imagen Video \cite{3-ho2022imagenvideohighdefinition} uses seven submodels in a cascade system. Recent work integrates LLMs: VideoDirectorGPT \cite{4-lin2023videodirectorgpt} for multiscene planning and Dysen-VDM \cite{5-fei2023empowering} for scene graphs.
In an unconditional generation, VIDM \cite{6-mei2023vidm} established a two-stream architecture. VDT \cite{7-lu2023vdt} introduced Transformers for spatio-temporal modeling, and VITRON \cite{8-fei2024vitron} combines LLMs with diffusion models.

\subsection{Pose-guided Character Motion Transfer}
Pose-Guided Character Motion Transfer synthesizes target character videos from pose signals, with research evolving from traditional explicit methods to modern diffusion-based approaches \cite{14-tu2024stableanimator,15-tan2024animateX,16-li2024dispose,17-tu2024motioneditor,18-wang2025humandreamer,tu2023implicit,tu2024motionfollower,tu2025stableanimator++,tu2025stableavatar}. Early techniques relying on optical flow estimation and Thin-Plate Spline transformations with GAN inpainting frequently exhibited artifacts, including texture tearing and temporal flickering \cite{9-chan2019everybody,10-zhu2023human}. Contemporary diffusion models have demonstrated superior performance in motion accuracy and detail preservation.
Recent innovations include MagicAnimate's \cite{11-xu2024magicanimate} integration of DensePose with temporal attention modules for effective motion-identity decoupling and Animate Anyone's \cite{12-hu2024animate_anyone} ReferenceNet combined with pose guidance for stable generation from single images. Champ \cite{13-zhu2024champ} advances the field by incorporating SMPL models with multimodal 3D signals (depth, normal, and semantic maps) to enhance alignment precision, while StableAnimator \cite{14-tu2024stableanimator} tackles identity drift through specialized facial preservation techniques.
Despite these advancements, current methods remain limited to mechanical motion transfer without capturing personalized characteristics. This fundamental constraint results in rigid, expressionless animations that significantly hinder applications in film production and immersive digital environments, where nuanced character expressiveness is crucial.

\subsection{Motion Style Transfer}

Motion Style Transfer methods \cite{zhong12024smoodi,zhong3li2024mulsmo,zhong4xie2024omnicontrol} transfer stylistic features while preserving target motion content integrity. Recent advances show diversified approaches and improved results. Aerman \textit{et al.} \cite{19-aberman2020unpaired} pioneer unpaired motion style transfer, extracting styles directly from videos for 3D animation. Finestyle \cite{20-song2023finestyle} enhances control through semantic-aware DIFF modules, while MCM-LDM \cite{21-song2024arbitrary} achieves Arbitrary Motion Style Transfer via trajectory-content-style decoupling. Most \cite{22-kim2024most} enables cross-content transfer using part-aware modulators, and PersonaBooth \cite{23-kim2025personabooth} introduces text-guided personalized generation. However, current methods rely on motion capture or SMPL data \cite{24-loper2023smpl}, which limits applications for special subjects, creating demand for more flexible solutions.

\section{METHODOLOGY}

\subsection{Pose Representation}
We categorize the input motion videos of the PersonaAnimator framework into two types based on function: content motion video $\bm{V}_c \in \mathbb{R}^{F \times 3 \times H \times W}$, which provides basic motion content (\emph{e.g.}, dancing, walking), and style motion video $\bm{V}_s \in \mathbb{R}^{F \times 3 \times H \times W}$, which offers personalized motion characteristics (\emph{e.g.}, gait, wave amplitude). Here, $c$ denotes motion content, $s$ denotes motion style, $F$ is the frame count, and $H$ and $W$ are the height and width of the video frames, respectively. Using DWpose \cite{dwpose}, we perform pose estimation to obtain the content skeleton pose sequence $\bm{P}_{c}^m \in \mathbb{R}^{F \times J \times 2}$ and the style skeleton pose sequence $\bm{P}_s^n \in \mathbb{R}^{F \times J \times 2}$, where $m$ and $n$ denote the pose content of the content and style poses, $J=20$ indicates the number of joints, and 2 represents the $x,y$ coordinates of each joint. PersonaAnimator aims to learn personalized features while preserving the basic motion structure of the pose content $m$, thereby generating a personalized pose $\bm{P}_s^m \in \mathbb{R}^{F \times J \times 2}$ that combines both characteristics.

\begin{figure*}[t]
  \centering
  \includegraphics[width=1.0\linewidth]{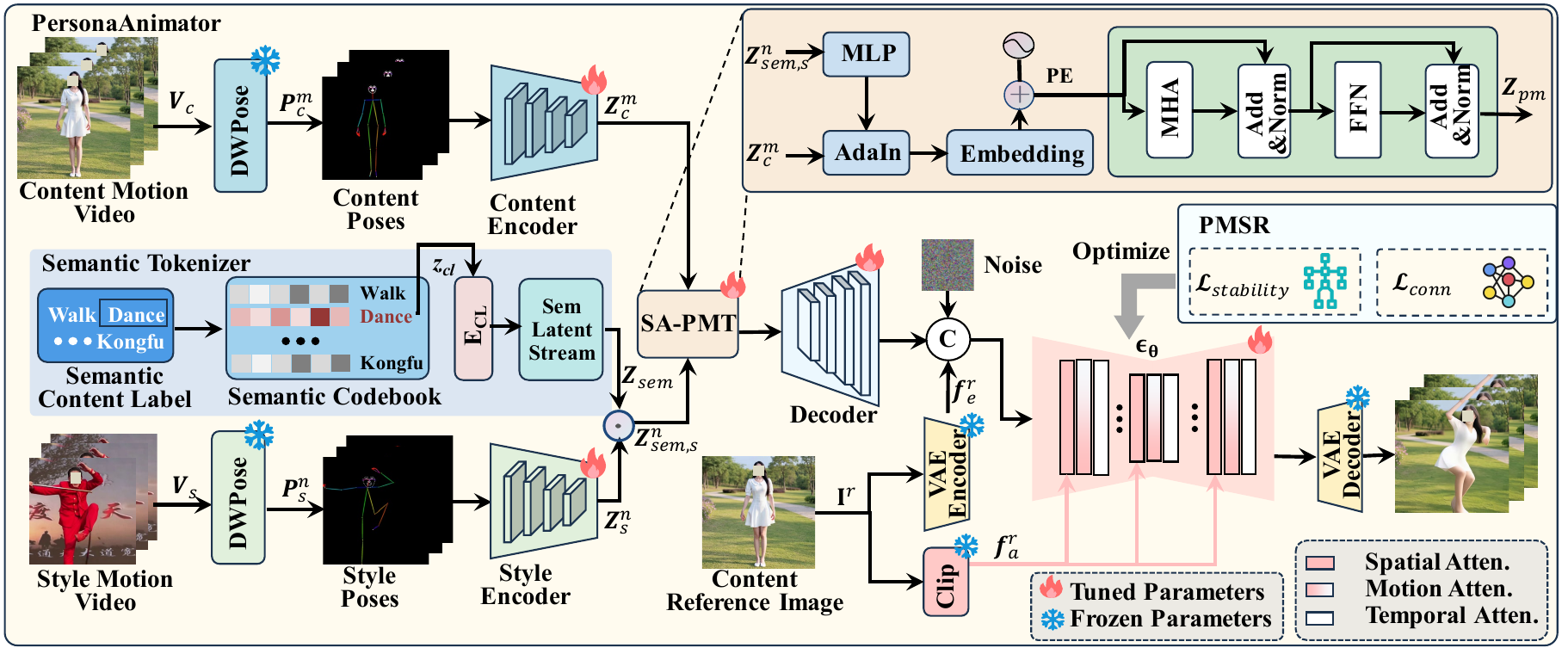}
  \caption{Architecture overview of PersonaAnimator framework. $\odot$ denotes the Hadamard product. $\oplus$ denotes the element-wise sum. $C$ denotes the concatenation operation.
  }
\label{Framework图}
\end{figure*}

\subsection{PersonaAnimator Framework Overview}
Prior work in computer vision has achieved basic motion generation, but key limitations remain:
 (1) Pose-guided character motion transfer methods \cite{15-tan2024animateX,14-tu2024stableanimator,16-li2024dispose} replicate motion mechanically without capturing personalized traits, severely limiting the expressiveness of digital characters.
 (2) Motion style transfer techniques \cite{22-kim2024most,23-kim2025personabooth}. However, capable of modeling style, relies fully or partially on motion capture data, which is costly, labor-intensive, and often unavailable for historical figures or virtual characters.
To address these issues, we propose a new task, Video-to-Video Motion Personalization, and introduce the PersonaAnimator framework, which requires no motion capture data and enables personalized motion transfer from unconstrained videos. This is critical for enhancing the realism of digital characters, as illustrated in Fig. \ref{Framework图}.

We first use DWPose \cite{dwpose} to extract skeleton poses $\bm{P}_c^m$  and $\bm{P}_s^n$  from content video $\bm{V}_c$ and style video $\bm{V}_s$, respectively. These poses are encoded into content features $\bm{Z}_c^m$  and style features $\bm{Z}_s^n$ via dedicated encoders. Meanwhile, semantic content labels are tokenized into $\bm{Z}_{\text {sem }}$, which performs Hadamard product with $\bm{Z}_s^n$ to produce $\bm{Z}_{\text {sem,s }}^n$. This semantic-guided weighting enhances motion-relevant style features while suppressing those that are irrelevant. The appendix provides a detailed description of the semantic tokenizer.

In Section SA-PMT Module, we propose the Semantic-Aware Personalized Motion Transfer (SA-PMT) Module, extracting basic motion content and personalized characteristics under semantic guidance and then fuses them into $\bm{Z}_{p m}$ features to generate personalized poses $\bm{P}_s^m$.

In the personalized motion generation process, we first extract a content reference frame $\bm{I}^{r}$ from $\bm{V}_c$ and encode it using pre-trained CLIP \cite{clip} and VAE \cite{vae} encoders to obtain appearance feature $\bm{f}_a^r$ and latent feature $\bm{f}_{e}^{r}$ respectively. These features are concatenated with personalized motion characteristics $\bm{Z}_{pm}$ and noise latent $\bm{Z}_{t}$ as input to the video diffusion model $\epsilon_\theta$. The model's spatial, motion, and temporal attention layers jointly capture human joint topology, motion dynamics, and inter-frame dependencies. The final personalized motion video $\bm{V}_s^m$ is generated through a VAE decoder \cite{vae}. In addition, we introduce the PMSR mechanism to enforce physically plausible personalized motion generation.

PersonaAnimator redefines motion generation by learning personalized motion features directly from videos, eliminating reliance on motion capture. Its decoupled architecture precisely preserves content motion structure while transferring personalized characteristics, establishing a novel ``postural approximation-to-personalized dynamic essence'' generation standard that sets a new benchmark for digital character animation.

\subsection{ Semantic-Aware Personalized Motion Transfer}
\label{sec:SA-PMT module}

Existing pose-guided motion transfer methods \cite{15-tan2024animateX,14-tu2024stableanimator} lack personalization, as they perform only rigid retargeting. Our Semantic-Aware Personalized Motion Transfer (SA-PMT) module learns both basic motion structures and personalized traits under semantic guidance, generating motions that preserve the advantages of both sources.
Specifically, we use AdaIN \cite{adain} to align content and style poses statistically, enabling effective feature fusion to produce the personalized feature $\bm{Z}_{ada}$:

\begin{equation}
\bm{Z}_{ada}=\text{AdaIN}\left(\bm{Z}_c^m, \text{MLP}\left(\bm{Z}_{sem,s}^n\right)\right),
\end{equation}
where $\bm{Z}_{c}^{m}$ is the content skeleton pose feature, while $\bm{Z}_{sem, s}^n$ denotes the semantic-conditioned personalized feature, with $\text{AdaIN}(\cdot)$ indicating the AdaIN structure. Subsequently, we encode $\bm{Z}_{ada}$, projecting it into a stable feature distribution space and incorporating positional encoding to preserve spatiotemporal information:

\begin{equation}
\bm{Z}_{p m}^c=\mathcal{E}_{SA-PMT}\left(\text{PE}\left(\bm{Z}_{ada}\right)\right),
\end{equation}
where $\mathcal{E}_{SA-PMT}(\cdot)$ denotes the encoder of the SA-PMT Module, primarily composed of fully-connected layers, while $\text{PE}(\cdot)$ represents positional encoding. We employ a Multi-Head Attention (MHA) mechanism \cite{transformer} to concurrently capture content-style relationships across different dimensions, yielding the final personalized motion feature vector $\bm{Z}_{pm}$, computed as:

\begin{equation}
\begin{gathered}
\bm{Z}_{pm}^f=\text{LN}\left(\bm{Z}_{pm}^c+\text{MHA}\left(\bm{Z}_{pm}^c\right)\right), \\
\bm{Z}_{pm}=\text{LN}\left(\bm{Z}_{p m}^f+\mathcal{F}\left(\bm{Z}_{pm}^f\right)\right),
\end{gathered}
\end{equation}
where $\text{LN}(\cdot)$ denotes Layer Normalization, $\mathcal{F}(\cdot)$. is the Feed-Forward Network, and $\bm{Z}_{pm}^c$, $\bm{Z}_{pm}^f$ are the module's personalized motion latent codes. The SA-PMT Module fuses features under semantic guidance, preserving motion content while learning subject-specific characteristics to generate personalized motions, thereby overcoming the limitations of current methods in rigid retargeting.

\subsection{Physics-aware Motion Style Regularization}
\label{sec:PMSR}

Current motion style transfer methods often struggle to simultaneously preserve physical plausibility and structural topology, leading to motion distortion. Prior works \cite{22-kim2024most,21-song2024arbitrary} mitigate this issue to some extent by introducing velocity and foot contact constraints. However, they lack explicit physical constraints on the full-body skeleton and ignore repulsive supervision between non-connected joints, resulting in physically implausible motions.
To address this, we propose the Physics-aware Motion Style Regularization (PMSR) mechanism, which enforces physically plausible motion generation through dynamic bone stability and body connectivity constraints.

\noindent \textbf{Dynamic Bone Stability Constraints.}
We penalize the second-order derivative (acceleration) of bone length variations to prevent unnatural stretching and maintain dynamic stability. Here, $\bm{p}_{f,j} \in \mathbb{R}^2$ denotes the 2D coordinates of joint $j$ in personalized pose $\bm{P}_s^m \in \mathbb{R}^{F \times J \times 2}$ at frame $f$. Let $\mathcal{B}=\left\{\left(i_k, j_k\right)\right\}_{k=1}^N$ represent the bone set, where $\left(i_k, j_k\right)$ indicates the indices of two connected joints for the $k$-th bone, and $N$ is the total bone count. For any given bone $(i, j) \in \mathcal{B}$, its length at frame $f$ is defined as:
\begin{equation}
\bm{l}_{f,(i, j)}=\left\|\bm{p}_{f, i}-\bm{p}_{f, j}\right\|_2.
\end{equation}

Here, $\|\cdot\|_2$ is the Euclidean norm. We penalize the temporal second-order difference of bone length sequence $\bm{l}_{f,(i, j)}$,  representing length variation acceleration. For frame $\bm{f} \in[2, \bm{F}-1]$, the second-order difference of bone $(i, j)$'s length is approximated as:
\vspace{-0.2cm}
\begin{equation}
\Delta^2 \bm{l}_{f,(i, j)}=\bm{l}_{f+1,(i, j)}-2 \bm{l}_{f,(i, j)}+\bm{l}_{f-1,(i, j)}
\end{equation}

Finally, we implement the dynamic bone stability constraint using the L2 norm, formulated as:
\begin{equation}
\mathcal{L}_{\text {stability}}=\frac{1}{\bm{F}-2} \sum_{f=2}^{F-1}\left\|\Delta^2 l_{f,(i, j)}\right\|_2^2.
\end{equation}

\noindent \textbf{Body Connectivity Constraints.} To preserve correct joint connectivity, we propose body connectivity constraints based on the adjacency matrix, which enhances the cohesion of connected joints while suppressing false proximity between unconnected joints. We define an adjacency matrix $\bm{A} \in\{0,1\}^{J \times J}$ where $\bm{A}_{i j}=1$ indicates whether joint $i$ and $j$ should be connected. For frame $f$, we construct a Euclidean distance matrix $\bm{D}^{(f)} \in \mathbb{R}^{J \times J}, \bm{D}^{(f)}(i, j) \leftarrow \bm{l}_{f,(i, j)}$. To strengthen joint aggregation, we minimize distances between connected joints as follows:

\begin{equation}
\mathcal{L}_{\text {conn }}^{+}=\frac{1}{\|\bm{~A}\|_0 \times \bm{F}} \sum_{\mathrm{f}=1}^{\bm{F}} \sum_{\mathrm{i}, \mathrm{j}} \bm{~A}_{\mathrm{ij}} \cdot \bm{D}^{(\mathrm{f})}(\mathrm{i}, \mathrm{j}).
\end{equation}

Meanwhile, to prevent unconnected joints from being too close to each other, we enforce a minimum distance constraint between them:
\begin{equation}
\small
\mathcal{L}_{\text {conn }}^{-}=\frac{1}{\|\bm{A}\|_0 \times F} \sum_{f=1}^F \sum_{i, j}\left(1-\bm{A}_{i j}\right) \cdot \max \left(0, \delta-\bm{D}^{(f)}(i, j)\right).
\end{equation}

The complete body connectivity constraints, with $\delta=0.1$ as the minimum distance threshold, are formulated as:
\begin{equation}
\mathcal{L}_{\text {conn }}=\mathcal{L}_{\text {conn }}^{+}+\mathcal{L}_{\text {conn }}^{-}.
\end{equation}

\noindent\textbf{Optimization Objectives.}
In addition, we adopt the $\mathcal{L}_{sc}$ proposed by Aberman \textit{et al.} \cite{19-aberman2020unpaired} to enforce style consistency, and the mean squared error (MSE) loss $\mathcal{L}_d$ used in prior works \cite{15-tan2024animateX} for appearance alignment and motion modeling. Further details are provided in the appendix. The final training loss is defined as:
\begin{equation}
\mathcal{L}_{\text {total }}=\mathcal{L}_{\text {stability }}+\mathcal{L}_{\text {conn }}+\mathcal{L}_{sc}+\mathcal{L}_d.
\end{equation}

\section{PersonaVid Dataset}
Existing datasets for the Video-to-Video Motion Personalization task exhibit notable limitations. Current motion capture datasets \cite{19-aberman2020unpaired,100style_dataset,xia2015realtime_dataset} can train motion style transfer models but rely on professional equipment and controlled environments, rendering data collection costly and impractical for deceased individuals or fictional characters. Pose-guided character motion transfer datasets \cite{tiktok_dataset,fashion_dataset} contain numerous videos but often lack precise style and content labels, which limits their direct applicability. Many videos also present low-quality issues, such as ambiguous motion styles and severe occlusions, further constraining their utility.

\begin{figure}[t]
  \centering
  \includegraphics[width=1\linewidth]{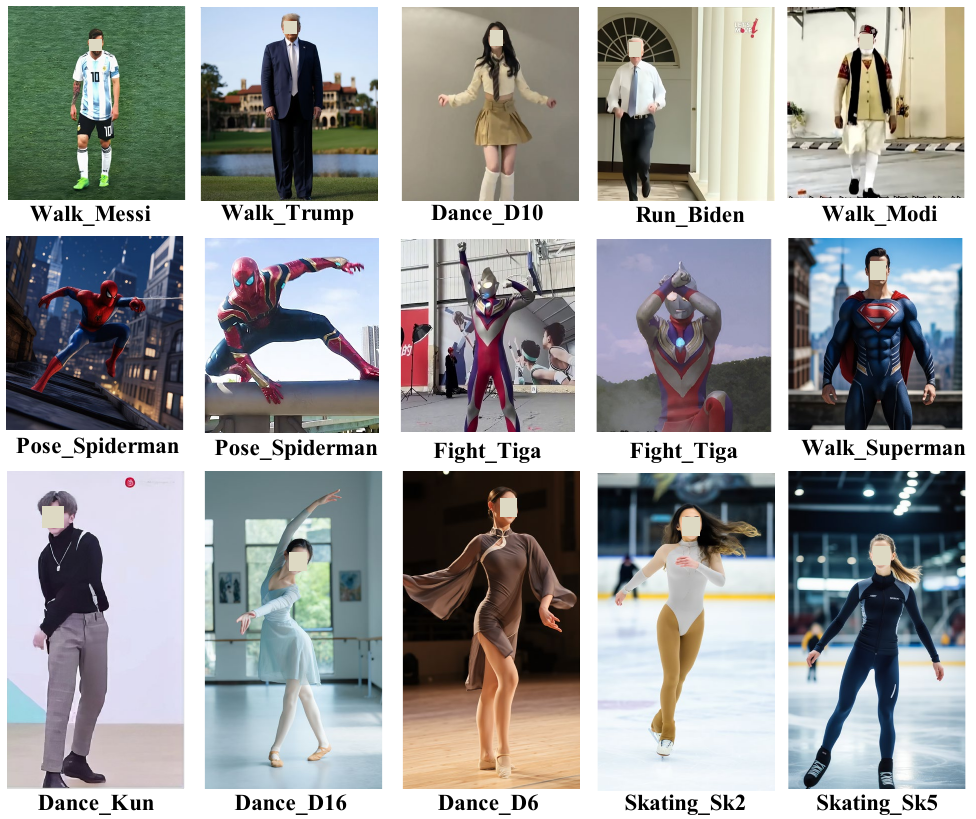}
  \caption{Examples of the PersonaVid dataset.
  }
  \vspace{-8pt}
\label{数据集示例图}
\end{figure}

To address this, we propose PersonaVid, the first video dataset for personalized motion learning. Constructed by collecting and annotating internet-sourced videos, it comprises 20 motion categories (\emph{e.g.}, walking, running) and 120 distinct styles. It covers a diverse range of performers, including ordinary individuals, athletes, and animated characters, thereby enhancing diversity in both individual variation and style representation. Table \ref{数据集比较} compares PersonaVid with existing datasets, while Figure \ref{数据集示例图} shows samples.


Current motion style transfer datasets \cite{xia2015realtime_dataset,19-aberman2020unpaired} use emotion categories (\emph{e.g.}, angry, depressed) as style labels, which misses individual specificity and lacks person-centric annotations. PersonaVid addresses this with a novel ``one-person-one-style'' paradigm, treating each individual as a unique style category to better capture subject-specific motion characteristics. Videos are annotated as ``Content\_Style\_Number,'' where ``Content'' denotes the basic motion (\emph{e.g.}, walking, running), and ``Number'' represents the video serial number. The ``Style'' is named in two ways: (1) If the video features famous figures, ``Style'' is abbreviated as their name (\emph{e.g.}, ``Walk\_Trump\_05'' for the fifth video of Trump walking). (2) For anonymous ordinary individuals, ``Style'' is denoted by the motion content abbreviation followed by a number (\emph{e.g.}, ``Dance\_D10\_01'' where ``D10'' indicates the style of the tenth anonymous dancer, or ``Skating\_Sk3\_01'' where ``Sk3'' represents the style of the third anonymous skater). The complete dataset taxonomy is detailed in the supplementary material.

    \begin{table}[t]
        \centering

        \setlength{\tabcolsep}{1pt}
        \renewcommand{\arraystretch}{1.2}
        \resizebox{\columnwidth}{!}{
        \begin{tabular}{ccccccc}
        \toprule
        \textbf{Dataset} & \textbf{Styles}   & \textbf{Contents}  & \textbf{MP4} & \textbf{Actor}   & \textbf{Frames Number}  & \textbf{PM} \\
        \midrule
        Xia \cite{xia2015realtime_dataset}  & 8          & 6   & 
        \textcolor{red}{\ding{55}}  & -          & 572  & \textcolor{red}{\ding{55}}     \\
        BFA \cite{19-aberman2020unpaired}        & 16     & 9     & \textcolor{red}{\ding{55}}          & 1   &32 &\textcolor{red}{\ding{55}}              \\
        BN-1 \cite{BN1-dataset}       & 15     & 17     & \textcolor{red}{\ding{55}} &2 &175           & \textcolor{red}{\ding{55}}      \\
        BN-2 \cite{BN1-dataset}       & 7          & 10          & \textcolor{red}{\ding{55}} &2 &2902    & \textcolor{red}{\ding{55}} \\
        100Style \cite{100style_dataset}       & 100          & 8          & \textcolor{red}{\ding{55}} &1 &810    & \textcolor{red}{\ding{55}} \\
        PerMo \cite{23-kim2025personabooth}       & 34          & 10          & \textcolor{red}{\ding{55}} &5 &6610    & \textcolor{red}{\ding{55}} \\        
        Tiktok \cite{tiktok_dataset}       & 0          & 0          & \textcolor{green}{\ding{51}} &- &92961    & \textcolor{red}{\ding{55}} \\
        Fashion \cite{fashion_dataset}       & 0          & 0          & \textcolor{green}{\ding{51}} &- &268474    & \textcolor{red}{\ding{55}}
        \\ \midrule
        \textbf{PersonaVid (Ours)}            & \textbf{120} & \textbf{20} & \textcolor{green}{\ding{51}} & \textbf{120} & 18867 &  \textcolor{green}{\ding{51}}               \\
        \bottomrule
        \end{tabular}
        }

    \caption{Comparison with existing datasets. ``PM'' denotes whether the dataset contains personalized motion, while ``-'' indicates no information.}
    \vspace{-8pt}
    \label{数据集比较}
    \end{table}

\section{Experiments}

\subsection{Implementation}
\noindent\textbf{Implementation Details.} 
We train and evaluate our model on the PersonaVid dataset. Training is conducted on an NVIDIA A800 using the RMSprop optimizer \cite{RMSprop_opt} (lr=le-4).

\noindent\textbf{Baseline Models.}
As no prior work addresses the Video-to-Video Motion Personalization task, we compare our PersonaAnimator framework with SOTA methods \cite{12-hu2024animate_anyone,14-tu2024stableanimator,15-tan2024animateX} in pose-guided motion transfer to show its superiority. However, these methods cannot learn personalized features from video motion. To ensure a fair comparison, we modify the three baseline frameworks as follows: 
We first equip them with the same content encoder, style encoder, and semantic tokenizer as used in PersonaAnimator, enabling them to take content and style videos as input.
 Next, we incorporate the Dual Interactive Flow Fusion (DIFF) module from FineStyle \cite{20-song2023finestyle} to allow the baselines to learn personalized features. The DIFF module has a similar parameter size to our SA-PMT module.
 Finally, the baselines utilize the DIFF-generated personalized motion for transfer, and we compare the results with those obtained using our method.
Detailed descriptions of the evaluation metrics and baseline models are provided in the appendix.

\begin{figure}[t]
  \centering
  \includegraphics[width=1\linewidth]{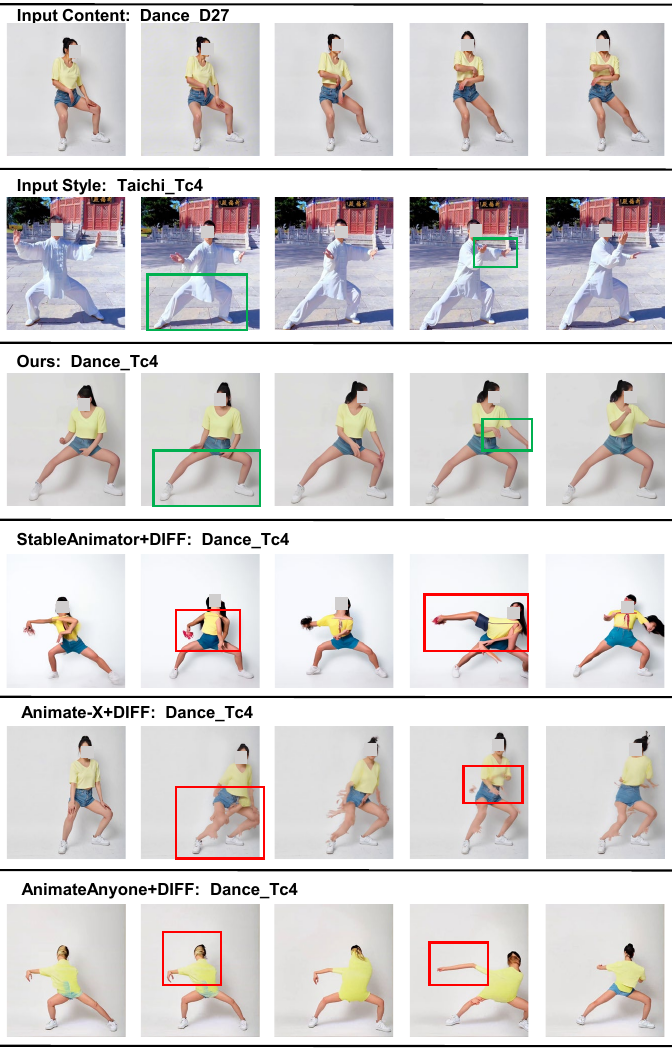}
  \caption{Qualitative comparison. Our framework outperforms SOTA methods, with green/red boxes marking learned/unlearned motion features, respectively. The results demonstrate the superior performance of our framework.
  }
\label{方法对比图1}
\end{figure}

\begin{figure}[t]
  \centering
  \includegraphics[width=1\linewidth]{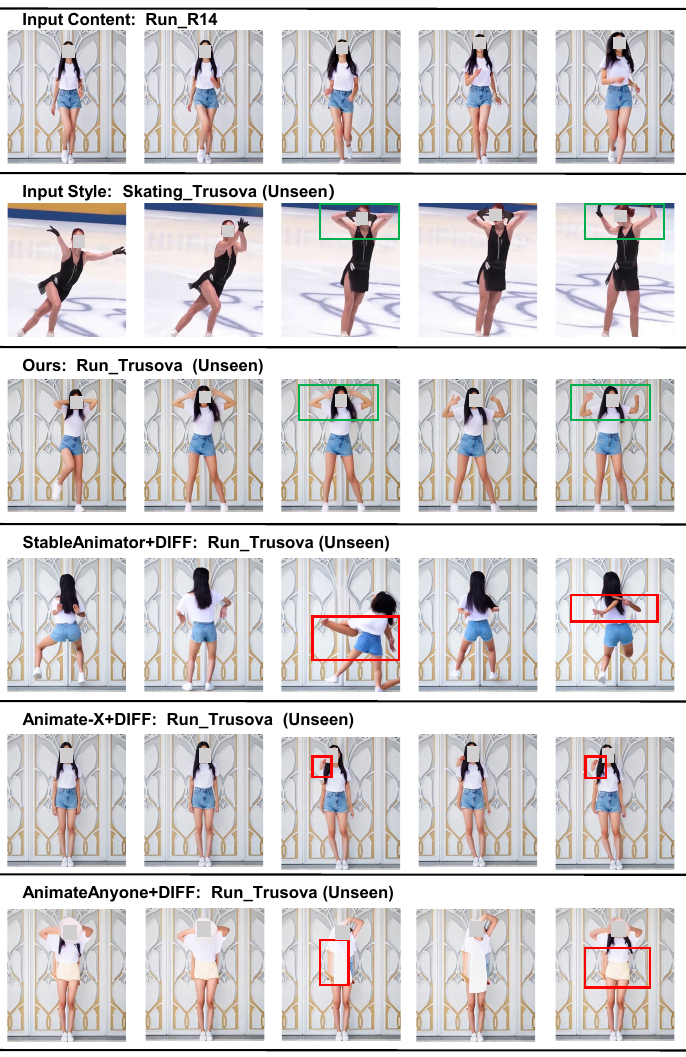}
  \caption{Generalization evaluation. We test different methods on an unseen style, with green/red boxes marking learned/unlearned motion features, respectively. Our method achieves better results than the SOTA methods.
  }
\label{unseen图}
\end{figure}

\subsection{Qualitative Evaluation}
We qualitatively compare PersonaAnimator with SOTA pose-guided motion transfer methods on style expressiveness and motion realism using the PersonaVid dataset. The model should preserve the basic motion content while learning personalized features to generate personalized motions, combining both aspects.

Visual comparison results are shown in Figure \ref{方法对比图1}, where our PersonaAnimator framework achieves superior generation quality. We use a dancing video of a female as the content motion and a Taichi video of an anonymous individual as the style motion, aiming to learn this person’s personalized Taichi features.
Our method captures key style traits, such as a lunge pose—where the left leg is bent and the right leg is extended—and both hands sweeping from left to right, as highlighted in the green box. Importantly, the motion is not mechanically copied but reflects a personalized adaptation.
In contrast, other methods produce distorted results:
StableAnimator+DIFF \cite{14-tu2024stableanimator} and Animate-X+DIFF \cite{15-tan2024animateX} capture the lunge pose but fail to reproduce hand movements accurately. AnimateAnyone+DIFF \cite{12-hu2024animate_anyone} generates a dancer facing away from the camera, showing only one hand.
None of the SOTA methods successfully learn the man’s personalized Taichi features.
These results stem from our SA-PMT module, which semantically guides the learning of motion features, and the PMSR mechanism, which enforces physical plausibility and prevents motion distortion.

\subsection{Generalizability}
Our framework extracts personalized features from arbitrary motions but may encounter unseen styles in practice. This generalization capability is crucial for practical applications.
To evaluate this, we select a skating video of Russian figure skater Trusova from the PersonaVid dataset, exclude it from training, and attempt to transfer her skating style to a running female subject. We apply the same procedure to SOTA methods and compare the generated results.


As shown in Figure \ref{unseen图}, even under unseen styles, our PersonaAnimator framework effectively captures Trusova’s personalized motion—crossed arms gradually opening—without mechanically copying the input (green box).In contrast, StableAnimator+DIFF \cite{14-tu2024stableanimator} and AnimateAnyone+DIFF \cite{12-hu2024animate_anyone} produce unnatural poses, while Animate-X+DIFF \cite{15-tan2024animateX} fails to learn any personalized characteristics. The generalizability comparison demonstrates that PersonaAnimator achieves superior generalization under unseen styles, enabling more realistic and vivid digital avatars in real-world scenarios.

\section{Conclusion}
This paper introduces the Video-to-Video Motion Personalization task, which enables personalized motion transfer from unconstrained videos, a crucial step in building identity-consistent digital avatars in the metaverse. To support this task, we introduce the large-scale PersonaVid dataset, which contains diverse motion content and style categories, along with the PersonaAnimator framework that achieves content-style decoupling and fusion through its SA-PMT module. Additionally, we design the PMSR mechanism to ensure physically plausible motion generation. Experimental results demonstrate that PersonaAnimator significantly outperforms existing methods, achieving SOTA performance in both qualitative and quantitative comparisons. The framework shows great potential for various applications, including virtual idol animation, film production, and game character animation generation.

\bibliography{aaai2026}

\end{document}